\documentclass{article} % For LaTeX2e
\usepackage{iclr2015,times}
\usepackage{hyperref}
\usepackage{url}

\usepackage{graphicx}
\usepackage{amsmath,amssymb}
\usepackage{array}
\usepackage{multirow}
\usepackage{pbox}
\usepackage{xcolor}
\usepackage{xspace}
\usepackage{nicefrac}
\usepackage[compact]{titlesec}

\title{Very Deep Convolutional Networks \\ for Large-Scale Image Recognition}

\author{
Karen Simonyan\thanks{current affiliation: Google DeepMind \; \textsuperscript{+}current affiliation: University of Oxford and Google DeepMind} \;
\& Andrew Zisserman\textsuperscript{+} \\
Visual Geometry Group, Department of Engineering Science, University of Oxford\\
\texttt{\{karen,az\}@robots.ox.ac.uk}  
}

\iclrfinalcopy
\iclrconference

\newcommand{\tblref}[1]{Table~\ref{#1}}
\newcommand{\sref}[1]{Sect.~\ref{#1}}
\newcommand{\apref}[1]{Appendix~\ref{#1}}

\newcommand*{\etal}{et~al.\@\xspace}
\newcommand*{\eg}{e.g.\@\xspace}
\newcommand*{\ie}{i.e.\@\xspace}
\newcommand*{\vs}{\emph{vs}.\@\xspace}

\begin{document}

\maketitle

\begin{abstract}
In this work we investigate the effect of the convolutional network depth on its accuracy in the large-scale image recognition setting.
Our main contribution is a thorough evaluation of networks of increasing depth using an architecture with very small ($3 \times 3$) convolution filters,
which shows that a significant improvement on the prior-art configurations can be achieved by pushing the depth to 16--19 weight layers. 
These findings were the basis of our \mbox{ImageNet} \mbox{Challenge} 2014 submission, where our team secured the first and the second places in the localisation and classification tracks
respectively. 
We also show that our representations generalise well to other datasets, where they achieve state-of-the-art results.
We have made our two best-performing ConvNet models publicly available to facilitate further research on the use of deep visual representations in computer vision.
\end{abstract}

\section{Introduction}
Convolutional networks (ConvNets) have recently enjoyed a great success in large-scale image and video recognition~\citep{Krizhevsky12,Zeiler13,Sermanet14,Simonyan14b}
which has become possible due to the large public image repositories, such as ImageNet~\citep{Deng09}, and high-performance computing systems, such as GPUs or large-scale distributed clusters~\citep{Dean12}.
In particular, an important role in the advance of deep visual recognition architectures has been played by the ImageNet Large-Scale Visual Recognition Challenge (ILSVRC)~\citep{Russakovsky14},
which has served as a testbed for a few generations of large-scale image classification systems, from high-dimensional shallow feature encodings~\citep{Perronnin10a} (the winner of ILSVRC-2011)
to deep ConvNets~\citep{Krizhevsky12} (the winner of ILSVRC-2012).

With ConvNets becoming more of a commodity in the computer vision field, a number of attempts have been made to improve the original architecture of~\citet{Krizhevsky12} in a bid to achieve
better accuracy. For instance, the best-performing submissions to the ILSVRC-2013~\citep{Zeiler13,Sermanet14} utilised smaller receptive window size and smaller stride of the first convolutional
layer. Another line of improvements dealt with training and testing the networks densely over the whole image and over multiple scales~\citep{Sermanet14,Howard14}.
In this paper, we address another important aspect of ConvNet architecture design -- its depth. 
To this end, we fix other parameters of the architecture, and steadily increase the depth of the network by adding more convolutional layers, which is feasible due to the use of very small ($3\times 3$) convolution filters in all layers.

As a result, we come up with significantly more accurate ConvNet architectures, which not only achieve the state-of-the-art accuracy on ILSVRC classification and localisation tasks, but are also applicable to other image recognition datasets, where they achieve excellent performance even when 
used as a part of a relatively simple pipelines (\eg deep features classified by a linear SVM without fine-tuning).
We have released our two best-performing models\footnote{\url{http://www.robots.ox.ac.uk/~vgg/research/very_deep/}} to facilitate further research.

The rest of the paper is organised as follows. In~\sref{sec:arch_config}, we describe our ConvNet configurations.
The details of the image classification training and evaluation are then presented in~\sref{sec:learning}, and the configurations are compared on the ILSVRC classification task in~\sref{sec:exp}. \sref{sec:conclusion} concludes the paper.
For completeness, we also describe and assess our ILSVRC-2014 object localisation system in~\apref{sec:loc}, and discuss the generalisation of very deep features to
other datasets in~\apref{sec:dataset_transfer}. Finally,~\apref{sec:revisions} contains the list of major paper revisions.

\section{ConvNet Configurations}
\label{sec:arch_config}
To measure the improvement brought by the increased ConvNet depth in a fair setting, all our \mbox{ConvNet} layer configurations are designed using the same principles, inspired by~\citet{Ciresan11,Krizhevsky12}.
In this section, we first describe a generic layout of our \mbox{ConvNet} configurations (\sref{sec:arch}) and then detail the specific configurations used in the evaluation (\sref{sec:config}).
Our design choices are then discussed and compared to the prior art in~\sref{sec:discuss}.

\subsection{Architecture}
\label{sec:arch}
During training, the input to our ConvNets is a fixed-size $224\times224$ RGB image. The only pre-processing we do is subtracting the mean RGB value, computed on the training set, from each pixel.
The image is passed through a stack of convolutional (conv.) layers, where we use filters with a very small receptive field: $3\times 3$ (which is the smallest size to capture the notion of 
left/right, up/down, center).
In one of the configurations we also utilise $1\times 1$ convolution filters, which can be seen as a linear transformation of the input channels (followed by non-linearity).
The convolution stride is fixed to $1$ pixel; the spatial padding of conv.\ layer input is such that the spatial resolution is preserved after convolution, \ie
the padding is $1$ pixel for $3\times3$ conv.\ layers.
Spatial pooling is carried out by five max-pooling layers, which follow some of the conv.\ layers (not all the conv.\ layers are followed by max-pooling).
Max-pooling is performed over a $2\times 2$ pixel window, with stride $2$.
% , so in total the spatial resolution of the input image is decreased by a factor of $2^5=32$.

A stack of convolutional layers (which has a different depth in different architectures) is followed by three Fully-Connected (FC) layers: the first
two have 4096 channels each, the third performs 1000-way ILSVRC classification and thus contains 1000 channels (one for each class). The final layer is the soft-max layer.
The configuration of the fully connected layers is the same in all networks. 

All hidden layers are equipped with the rectification (ReLU~\citep{Krizhevsky12}) non-linearity. 
We note that none of our networks (except for one) contain Local Response Normalisation (LRN) normalisation~\citep{Krizhevsky12}: as will be shown in~\sref{sec:exp}, such normalisation 
does not improve the performance on the ILSVRC dataset, but leads to increased memory consumption and computation time. 
Where applicable, the parameters for the LRN layer are those of~\citep{Krizhevsky12}.

\subsection{Configurations}
\label{sec:config}
The ConvNet configurations, evaluated in this paper, are outlined in~\tblref{tab:config}, one per column. In the following we will refer to the nets by their names 
(A--E). All configurations follow the generic design presented in~\sref{sec:arch}, and differ only in the depth: 
from 11 weight layers in the network A (8 conv.\ and 3 FC layers) to 19 weight layers in the network E (16 conv.\ and 3 FC layers).
The width of conv.\ layers (the number of channels) is rather small, starting from $64$ in the first layer and then increasing by a factor of $2$ after each max-pooling layer,
until it reaches $512$.

In~\tblref{tab:num_params} we report the number of parameters for each configuration.
In spite of a large depth, the number of weights in our nets is not greater than the number of weights in a more shallow net with larger conv.\ layer widths and receptive fields
(144M weights in~\citep{Sermanet14}).

\begin{table}[htb]
\centering
\small
\caption{\textbf{ConvNet configurations} (shown in columns).
The depth of the configurations increases from the left (A) to the right (E), as more layers are added (the added layers are shown in bold).
The convolutional layer parameters are denoted as ``conv$\langle$receptive field size$\rangle$-$\langle$number of channels$\rangle$''.
The ReLU activation function is not shown for brevity.
}
\begin{tabular}{|c|c|c|c|c|c|} \hline
\multicolumn{6}{|c|}{{ConvNet Configuration}} \\ \hline
A & A-LRN & B & C & D & E \\ \hline
11 weight & 11 weight & 13 weight & 16 weight & 16 weight & 19 weight \\ 
layers & layers & layers & layers & layers & layers \\ \hline\hline
\multicolumn{6}{|c|}{input ($224 \times 224$ RGB image)} \\ \hline
conv3-64 & conv3-64 & conv3-64 & conv3-64 & conv3-64 & conv3-64 \\ 
 & \textbf{LRN} & \textbf{conv3-64} & conv3-64 & conv3-64 & conv3-64\\ \hline
\multicolumn{6}{|c|}{maxpool} \\ \hline
conv3-128 & conv3-128 & conv3-128 & conv3-128 & conv3-128 & conv3-128 \\ 
 & & \textbf{conv3-128} & conv3-128 & conv3-128 & conv3-128 \\ \hline
\multicolumn{6}{|c|}{maxpool} \\ \hline
conv3-256 & conv3-256 & conv3-256 & conv3-256 & conv3-256 & conv3-256 \\ 
conv3-256 & conv3-256 & conv3-256 & conv3-256 & conv3-256 & conv3-256 \\ 
& & & \textbf{conv1-256} & \textbf{conv3-256} & conv3-256 \\ 
& & & & & \textbf{conv3-256} \\ \hline
\multicolumn{6}{|c|}{maxpool} \\ \hline
conv3-512 & conv3-512 & conv3-512 & conv3-512 & conv3-512 & conv3-512 \\ 
conv3-512 & conv3-512 & conv3-512 & conv3-512 & conv3-512 & conv3-512 \\ 
& & & \textbf{conv1-512} & \textbf{conv3-512} & conv3-512 \\ 
& & & & & \textbf{conv3-512} \\ \hline
\multicolumn{6}{|c|}{maxpool} \\ \hline
conv3-512 & conv3-512 & conv3-512 & conv3-512 & conv3-512 & conv3-512 \\ 
conv3-512 & conv3-512 & conv3-512 & conv3-512 & conv3-512 & conv3-512 \\ 
& & & \textbf{conv1-512} & \textbf{conv3-512} & conv3-512 \\ 
& & & & & \textbf{conv3-512} \\ \hline
\multicolumn{6}{|c|}{maxpool} \\ \hline
\multicolumn{6}{|c|}{FC-4096} \\ \hline
\multicolumn{6}{|c|}{FC-4096} \\ \hline
\multicolumn{6}{|c|}{FC-1000} \\ \hline
\multicolumn{6}{|c|}{soft-max} \\ \hline
\end{tabular}
\label{tab:config}
\end{table}
\begin{table}[htb]
\small
\centering
\caption{\textbf{Number of parameters} (in millions).}
\begin{tabular}{|l|c|c|c|c|c|c|} \hline
Network & A,A-LRN & B & C & D & E \\ \hline
Number of parameters & 133 & 133 & 134 & 138 & 144 \\ \hline
\end{tabular}
\label{tab:num_params}
\end{table}

\subsection{Discussion}
\label{sec:discuss}
% Relation to other ConvNet architectures
Our ConvNet configurations are quite different from the ones used in the top-performing entries of the ILSVRC-2012~\citep{Krizhevsky12} and ILSVRC-2013 competitions~\citep{Zeiler13,Sermanet14}.
Rather than using relatively large receptive fields in the first conv.\ layers (\eg $11\times 11$ with stride $4$ in~\citep{Krizhevsky12}, or $7 \times 7$ with stride $2$ in~\citep{Zeiler13,Sermanet14}),
we use very small $3 \times 3$ receptive fields throughout the whole net, which are convolved with the input at every pixel (with stride $1$).
It is easy to see that a stack of two $3 \times 3$ conv.\ layers (without spatial pooling in between) has an effective receptive field of $5 \times 5$;
three such layers have a $7 \times 7$ effective receptive field. 
So what have we gained by using, for instance, a stack of three $3 \times 3$ conv.\ layers instead of a single $7 \times 7$ layer?
First, we incorporate three non-linear rectification layers instead of a single one, which makes the decision function more discriminative.
Second, we decrease the number of parameters: assuming that both the input and the output of a three-layer $3 \times 3$ convolution stack has $C$ channels, the stack is parametrised
by $3 \left( 3^2 C^2 \right) = 27 C^2$ weights; at the same time, a single $7\times 7$ conv.\ layer would require $7^2 C^2=49 C^2$ parameters, \ie $81\%$ more.
This can be seen as imposing a regularisation on the $7\times7$ conv.\ filters, forcing them to have a decomposition through the $3 \times 3$ filters (with non-linearity injected in between).
% \note{try to describe the decomposition formally, in terms of rank}.

The incorporation of $1\times 1$ conv.\ layers (configuration C,~\tblref{tab:config}) is a way to increase the non-linearity of the decision function without affecting
the receptive fields of the conv.\ layers. Even though in our case the $1\times 1$ convolution is essentially a linear projection onto the space of the same dimensionality 
(the number of input and output channels is the same), an additional non-linearity is introduced by the rectification function.
It should be noted that $1\times 1$ conv.\ layers have recently been utilised in the ``Network in Network'' architecture of~\citet{Lin14}.

Small-size convolution filters have been previously used by~\citet{Ciresan11}, but their nets are significantly less deep than ours, and they did not
evaluate on the large-scale ILSVRC dataset. 
\citet{Goodfellow13} applied deep ConvNets ($11$ weight layers) to the task of street number recognition, and showed that the increased depth led to better performance.
GoogLeNet~\citep{Szegedy14}, a top-performing entry of the ILSVRC-2014 classification task, was developed independently of our work, but is similar in that it is based on very deep ConvNets (22 weight layers) and small convolution filters (apart from $3\times 3$, they also use $1\times 1$ and \mbox{$5\times 5$} convolutions). Their network topology is, however, more complex than ours, and the spatial resolution of the feature maps is reduced more aggressively in the first layers to decrease the amount of computation. As will be shown in~\sref{sec:class_SOA}, our model is outperforming that of~\citet{Szegedy14} in terms of the single-network classification accuracy.

\section{Classification Framework}
\label{sec:learning}
In the previous section we presented the details of our network configurations. In this section, we describe the details of classification ConvNet training and evaluation.

\subsection{Training}
\label{sec:train}
The ConvNet training procedure generally follows~\citet{Krizhevsky12} (except for sampling the input crops from multi-scale training images, as explained later).
Namely, the training is carried out by optimising the multinomial logistic regression objective using mini-batch gradient descent (based on back-propagation~\citep{LeCun89}) with momentum. The batch size was set to $256$, momentum to $0.9$.
The training was regularised by weight decay (the $L_2$ penalty multiplier set to $5\cdot 10^{-4}$) and dropout regularisation for the first two fully-connected layers (dropout ratio set to $0.5$).
The learning rate was initially set to $10^{-2}$, and then decreased
by a factor of $10$ when the validation set accuracy stopped improving. In total, the learning rate was decreased 3 times, and the learning was stopped after $370$K iterations
(74 epochs). We conjecture that in spite of the larger number of parameters and the greater depth of our nets compared to~\citep{Krizhevsky12}, the nets required less epochs
to converge due to (a)~implicit regularisation imposed by greater depth and smaller conv.\ filter sizes; (b)~pre-initialisation of certain layers.

The initialisation of the network weights is important, since bad initialisation can stall learning due to the instability of gradient in deep nets.
To circumvent this problem, we began with training the configuration A (\tblref{tab:config}), shallow enough to be trained with random initialisation. Then, when training deeper
architectures, we initialised the first four convolutional layers and the last three fully-connected layers with the layers of net A (the intermediate layers were
initialised randomly). We did not decrease the learning rate for the pre-initialised layers, allowing them to change during learning.
For random initialisation (where applicable), we sampled the weights from a normal distribution with the zero mean and $10^{-2}$ variance. The biases were initialised with zero.
It is worth noting that after the paper submission we found that it is possible to initialise the weights without pre-training by using the random initialisation procedure of~\citet{Glorot10}.

To obtain the fixed-size $224 \times 224$ ConvNet input images, they were randomly cropped from rescaled training images (one crop per image per SGD iteration).
To further augment the training set, the crops underwent random horizontal flipping and random RGB colour shift~\citep{Krizhevsky12}. Training image rescaling is explained below.

\paragraph{Training image size.}
Let $S$ be the smallest side of an isotropically-rescaled training image, from which the ConvNet input is cropped (we also refer to $S$ as the training scale).
While the crop size is fixed to $224 \times 224$, in principle $S$ can take on any value not less than $224$: for $S=224$ the crop will capture whole-image statistics, completely spanning the smallest side of a training image;
for $S \gg 224$ the crop will correspond to a small part of the image, containing a small object or an object part.

We consider two approaches for setting the training scale $S$. 
The first is to fix $S$, which corresponds to single-scale training (note that image content within the sampled crops can still represent multi-scale image statistics).
In our experiments, we evaluated models trained at two fixed scales: $S=256$ (which has been widely used in the prior art~\citep{Krizhevsky12,Zeiler13,Sermanet14}) and $S=384$.
Given a \mbox{ConvNet} configuration, we first trained the network using $S=256$. 
% We then trained another network with the same layer configuration, but using $S=384$.
To speed-up training of the $S=384$ network, it was initialised with the weights pre-trained with $S=256$, and we used a smaller initial learning rate of $10^{-3}$.

The second approach to setting $S$ is multi-scale training, where each training image is individually rescaled by randomly sampling $S$ from a certain range $\left[S_{min},S_{max}\right]$ (we used $S_{min}=256$ and $S_{max}=512$). 
Since objects in images can be of different size, it is beneficial to take this into account during training.
This can also be seen as training set augmentation by scale jittering, where a single model is trained to recognise objects over a wide range of scales.
For speed reasons, we trained multi-scale models by fine-tuning all layers of a single-scale model with the same configuration, pre-trained with fixed $S=384$.

\subsection{Testing}
\label{sec:test}
At test time, given a trained ConvNet and an input image, it is classified in the following way. 
First, it is isotropically rescaled to a pre-defined smallest image side, denoted as $Q$ (we also refer to it as the test scale).
We note that $Q$ is not necessarily equal to the training scale $S$ (as we will show in~\sref{sec:exp},
using several values of $Q$ for each $S$ leads to improved performance).
Then, the network is applied densely over the rescaled test image in a way similar to~\citep{Sermanet14}. Namely, the fully-connected layers are first converted to convolutional layers (the first FC layer to a $7\times7$ conv.\ layer, the last two FC layers to $1\times1$ conv.\ layers). The resulting fully-convolutional net is then applied to the whole (uncropped) image.
% by convolving the filters in each layer with the full-size input.
The result is a class score map with the number of channels equal to the number of classes, and a variable spatial resolution, dependent
on the input image size. Finally, to obtain a fixed-size vector of class scores for the image, the class score map is spatially averaged (sum-pooled).
We also augment the test set by horizontal flipping of the images; the soft-max class posteriors of the original and flipped images are averaged to obtain
the final scores for the image.

Since the fully-convolutional network is applied over the whole image, there is no need to sample multiple crops at test time~\citep{Krizhevsky12}, which is less efficient as it requires network re-computation for each crop. At the same time, using a large set of crops, as done by~\citet{Szegedy14}, can lead to improved accuracy, as it results in a finer sampling of the input image compared to the fully-convolutional net. 
Also, multi-crop evaluation is complementary to dense evaluation due to different convolution boundary conditions: when applying a ConvNet to a crop, the convolved feature maps are padded with zeros, while 
in the case of dense evaluation the padding for the same crop naturally comes from the neighbouring parts of an image (due to both the convolutions and spatial pooling), which substantially increases the overall network receptive field, so more context is captured.
While we believe that in practice the increased computation time of multiple crops does not justify the potential gains in accuracy, for reference we also evaluate our networks using $50$ crops per scale ($5 \times 5$ regular grid with $2$ flips), for a total of $150$ crops over $3$ scales, which is comparable to $144$ crops over $4$ scales used by~\citet{Szegedy14}.

\subsection{Implementation Details}
Our implementation is derived from the publicly available C++ Caffe toolbox~\citep{Jia13} (branched out in December 2013), but contains a number of significant modifications, 
allowing us to perform training and evaluation on multiple GPUs installed in a single system, as well as train and evaluate on full-size (uncropped) images at multiple scales (as described above).
Multi-GPU training exploits data parallelism, and is carried out by splitting each batch of training images into several GPU batches, processed in parallel on each GPU. 
After the GPU batch gradients are computed, they are averaged to obtain the gradient of the full batch.
Gradient computation is synchronous across the GPUs, so the result is exactly the same as when training on a single GPU.

While more sophisticated methods of speeding up ConvNet training have been recently proposed~\citep{Krizhevsky14}, which employ model and data parallelism for different layers
of the net, we have found that our conceptually much simpler scheme already provides a speedup of $3.75$ times on an off-the-shelf \mbox{4-GPU} system, as compared to using a single GPU.
On a system equipped with four NVIDIA Titan Black GPUs, training a single net took 2--3 weeks depending on the architecture.

\section{Classification Experiments}
\label{sec:exp}

\paragraph{Dataset.}
In this section, we present the image classification results achieved by the described ConvNet architectures on the ILSVRC-2012 dataset (which was used for ILSVRC 2012--2014 challenges).
The dataset includes images of 1000 classes, and is split into three sets: training ($1.3$M images), validation ($50$K images), and testing ($100$K images with held-out class labels).
The classification performance is evaluated using two measures: the top-1 and top-5 error. The former is a multi-class classification error, \ie the proportion of incorrectly
classified images; the latter is the main evaluation criterion used in ILSVRC, and is computed as the proportion of images such that the ground-truth category is outside the top-5 predicted categories.

For the majority of experiments, we used the validation set as the test set. Certain experiments were also carried out on the test set and submitted to the official ILSVRC server 
as a ``VGG'' team entry to the ILSVRC-2014 competition~\citep{Russakovsky14}.

\subsection{Single Scale Evaluation}

We begin with evaluating the performance of individual ConvNet models at a single scale with the layer configurations described in~\sref{sec:config}.
The test image size was set as follows: $Q=S$ for fixed $S$, and $Q=0.5(S_{min}+S_{max})$ for jittered $S \in \left[S_{min},S_{max}\right]$.
The results of are shown in~\tblref{tab:results_single_scale}.

First, we note that using local response normalisation (A-LRN network) does not improve on the model A
without any normalisation layers. We thus do not employ normalisation in the deeper architectures (B--E).

Second, we observe that the classification error decreases with the increased ConvNet depth: from 11 layers in A to 19 layers in E.
Notably, in spite of the same depth, the configuration C (which contains three $1 \times 1$ conv.\ layers), performs worse than
the configuration D, which uses $3 \times 3$ conv.\ layers throughout the network.
This indicates that while the additional non-linearity does help (C is better than B), it is also important to capture spatial context by using conv.\ filters with non-trivial receptive fields (D is better than C). 
The error rate of our architecture saturates when the depth reaches $19$ layers, but even deeper models might be beneficial for larger datasets.
We also compared the net B with a shallow net with five $5\times5$ conv. layers, which was derived from B by replacing 
each pair of $3\times3$ conv. layers with a single $5\times5$ conv. layer (which has the same receptive field as explained in~\sref{sec:discuss}).
The top-1 error of the shallow net was measured to be $7\%$ higher than that of B (on a center crop), which confirms that a deep net with small 
filters outperforms a shallow net with larger filters. 

Finally, scale jittering at training time ($S \in [256;512]$) leads to significantly better results than training on images with fixed smallest side ($S=256$ or $S=384$),
even though a single scale is used at test time. This confirms that training set augmentation by scale jittering is indeed helpful for capturing multi-scale image statistics.

% It should also be noted that the ConvNets trained on larger $384 \times N$ images ($S=384$,~\sref{sec:train}) perform better than the nets trained on
% $256 \times N$ images. A possible reason could be that $384 \times N$ nets were obtained by fine-tuning $256 \times N$ nets (\sref{sec:learning}), which
% effectively increases the amount of training data. 
% Additionally, dense ConvNet computation over larger $384 \times N$ test images results in higher spatial resolution of the class score map (\sref{sec:test}), so the
% final (sum-pooled) class scores accumulate a larger numbers of decisions.
% We did not notice any performance improvement when moving further to $512 \times N$ images (results are not reported), which can be explained that 224x224 window in a 512xN can be
% too small to capture the object.

\vspace{-2em}

%%%%%%%%%%%%%%%%%%%%%%%%%%%%%%%%%%%%%
\begin{table}[htb]
\small
\centering
\caption{\textbf{ConvNet performance at a single test scale.}
}
\begin{tabular}{|l|c|c|c|c|} \hline
ConvNet config.\ (\tblref{tab:config}) & \multicolumn{2}{c|}{smallest image side} & top-1 val.\ error (\%) & top-5 val.\ error (\%) \\ \cline{2-3}
& train ($S$) & test ($Q$) & & \\ \hline
A & 256 & 256 & 29.6 & 10.4  \\ \hline
A-LRN & 256 & 256 & 29.7 & 10.5  \\ \hline
B & 256 & 256 & 28.7 & 9.9  \\ \hline
\multirow{3}{*}{C} & 256 & 256 & 28.1 & 9.4  \\ \cline{2-5}
 & 384 & 384 & 28.1 & 9.3  \\ \cline{2-5}
 & [256;512] & 384 & 27.3  & 8.8  \\ \hline
 \multirow{3}{*}{D} & 256 & 256 & 27.0  & 8.8  \\ \cline{2-5}
 & 384 & 384 & 26.8  & 8.7  \\ \cline{2-5}
 & [256;512] & 384 & 25.6  & 8.1  \\ \hline
 \multirow{3}{*}{E} & 256 & 256 & 27.3  & 9.0  \\ \cline{2-5}
 & 384 & 384 & 26.9  & 8.7  \\ \cline{2-5}
 & [256;512] & 384 & \textbf{25.5}  & \textbf{8.0}  \\ \hline
\end{tabular}
\label{tab:results_single_scale}
\end{table}
%%%%%%%%%%%%%%%%%%%%%%%%%%%%%%%%%%%%%

\subsection{Multi-Scale Evaluation}
\label{sec:multi-scale}

Having evaluated the ConvNet models at a single scale, we now assess the effect of scale jittering at test time.
It consists of running a model over several rescaled versions of a test image (corresponding to different values of $Q$), followed by averaging the resulting class posteriors.
Considering that a large discrepancy between training and testing scales leads to a drop in performance, the models trained with fixed $S$ were evaluated over 
three test image sizes, close to the training one: $Q=\{S-32, S, S+32\}$. 
At the same time, scale jittering at training time allows the network to be applied to a wider range of scales at test time, so the model trained
with variable $S \in [S_{min}; S_{max}]$ was evaluated over a larger range of sizes $Q=\{S_{min}, 0.5(S_{min} + S_{max}), S_{max}\}$. 

The results, presented in~\tblref{tab:results_multi_scale}, indicate that scale jittering at test time leads to better performance (as compared to evaluating 
the same model at a single scale, shown in~\tblref{tab:results_single_scale}). 
As before, the deepest configurations (D and E) perform the best, and scale jittering is better than training with a fixed smallest side $S$.
Our best single-network performance on the validation set is $24.8\%/7.5\%$ top-1/top-5 error (highlighted in bold in~\tblref{tab:results_multi_scale}).
On the test set, the configuration E achieves $7.3\%$ top-5 error.

\vspace{-1em}

%%%%%%%%%%%%%%%%%%%%%%%%%%%%%%%%%%%%%
\begin{table}[htb]
\small
\centering
\caption{\textbf{ConvNet performance at multiple test scales.}
}
\begin{tabular}{|l|c|c|c|c|} \hline
ConvNet config.\ (\tblref{tab:config}) & \multicolumn{2}{c|}{smallest image side} & top-1 val.\ error (\%) & top-5 val.\ error (\%) \\ \cline{2-3}
& train ($S$) & test ($Q$) & & \\ \hline
B & 256 & 224,256,288 & 28.2 & 9.6  \\ \hline
\multirow{3}{*}{C} & 256 & 224,256,288 & 27.7 & 9.2  \\ \cline{2-5}
 & 384 & 352,384,416 & 27.8 & 9.2  \\ \cline{2-5}
 & $\left[256;512\right]$ & 256,384,512 & 26.3 & 8.2  \\ \hline 
\multirow{3}{*}{D} & 256 & 224,256,288 & 26.6  & 8.6  \\ \cline{2-5}
 & 384 & 352,384,416 & 26.5  & 8.6  \\ \cline{2-5}
 & $\left[256;512\right]$ & 256,384,512 & \textbf{24.8}  & \textbf{7.5}  \\ \hline 
\multirow{3}{*}{E} & 256 & 224,256,288 & 26.9  & 8.7  \\ \cline{2-5}
 & 384 & 352,384,416 & 26.7  & 8.6  \\ \cline{2-5}
 & $\left[256;512\right]$ & 256,384,512 & \textbf{24.8}  & \textbf{7.5}  \\ \hline 
\end{tabular}
\label{tab:results_multi_scale}
\end{table}
%%%%%%%%%%%%%%%%%%%%%%%%%%%%%%%%%%%%%

\subsection{Multi-crop evaluation}
\label{sec:multi-crop}
In~\tblref{tab:results_multi_crop} we compare dense ConvNet evaluation with mult-crop evaluation (see~\sref{sec:test} for details). We also assess the complementarity of the two evaluation techniques by averaging their soft-max outputs. As can be seen, using multiple crops performs slightly better than dense evaluation, and the two approaches are indeed complementary, as their combination outperforms each of them. As noted above, we hypothesize that this is due to a different treatment of convolution boundary conditions.

% \vspace{-1em}
%%%%%%%%%%%%%%%%%%%%%%%%%%%%%%%%%%%%%
\begin{table}[htb]
\small
\centering
\caption{\textbf{ConvNet evaluation techniques comparison.}
In all experiments the training scale $S$ was sampled from $\left[256;512\right]$, and three test scales $Q$ were considered: $\left\{256,384,512\right\}$.
}
\begin{tabular}{|l|c|c|c|} \hline
ConvNet config.\ (\tblref{tab:config}) & Evaluation method & top-1 val.\ error (\%) & top-5 val.\ error (\%) \\ \hline
\multirow{3}{*}{D} & dense & 24.8  & 7.5  \\ \cline{2-4}
& multi-crop & 24.6 & 7.5  \\ \cline{2-4}
& multi-crop \& dense & \textbf{24.4}  & \textbf{7.2}  \\ \hline 
\multirow{3}{*}{E} & dense & 24.8  & 7.5  \\ \cline{2-4}
& multi-crop & 24.6 & 7.4  \\ \cline{2-4}
& multi-crop \& dense & \textbf{24.4}  & \textbf{7.1}  \\ \hline 
\end{tabular}
\label{tab:results_multi_crop}
\end{table}
%%%%%%%%%%%%%%%%%%%%%%%%%%%%%%%%%%%%%

\subsection{ConvNet Fusion}
\label{sec:fusion}
Up until now, we evaluated the performance of individual ConvNet models.
In this part of the experiments, we combine the outputs of several models by averaging their soft-max class posteriors.
This improves the performance due to complementarity of the models, and was used in the top ILSVRC submissions in 2012~\citep{Krizhevsky12} and 2013~\citep{Zeiler13,Sermanet14}.

The results are shown in~\tblref{tab:results_fusion}.
By the time of ILSVRC submission we had only trained the single-scale networks, as well as a multi-scale model D (by fine-tuning only the fully-connected layers rather than all layers). The resulting ensemble of 7 networks has $7.3\%$ ILSVRC test error.
% For reference, in~\tblref{tab:results_fusion} (last row) we report the results of combining these models.
After the submission, we considered an ensemble of only two best-performing multi-scale models (configurations D and E), which reduced the test error to 
$7.0\%$ using dense evaluation and $6.8\%$ using combined dense and multi-crop evaluation.
For reference, our best-performing single model achieves $7.1\%$ error (model E,~\tblref{tab:results_multi_crop}).

%%%%%%%%%%%%%%%%%%%%%%%%%%%%%%%%%%%%%
\begin{table}[htb]
\setlength{\tabcolsep}{2pt}
\small
\centering
\caption{\textbf{Multiple ConvNet fusion results.}
% Combined models are denoted as \\ ``(configuration name/train image size/test image sizes)'' (see~\tblref{tab:results_multi_scale} for individual model results).
}
\begin{tabular}{|l|c|c|c|} \hline
\multirow{2}{*}{Combined ConvNet models} & \multicolumn{3}{c|}{Error} \\ \cline{2-4}
 & top-1 val & top-5 val & top-5 test \\ \hline
 \multicolumn{4}{|c|}{ILSVRC submission} \\ \hline
 \pbox{11cm}{\vspace{0.2em}
 (D/256/224,256,288), (D/384/352,384,416), (D/[256;512]/256,384,512) \\ (C/256/224,256,288), (C/384/352,384,416) \\ (E/256/224,256,288), (E/384/352,384,416)} & 24.7 & 7.5 & 7.3 \\ \hline
 \multicolumn{4}{|c|}{post-submission} \\ \hline
 \pbox{11cm}{\vspace{0.2em}
(D/[256;512]/256,384,512), (E/[256;512]/256,384,512)}, dense eval. & 24.0 & 7.1 & 7.0 \\ \hline
\pbox{11cm}{\vspace{0.2em}
(D/[256;512]/256,384,512), (E/[256;512]/256,384,512)}, multi-crop & 23.9 & 7.2 & - \\ \hline
\pbox{11cm}{\vspace{0.2em}
(D/[256;512]/256,384,512), (E/[256;512]/256,384,512)}, multi-crop \& dense eval. & \textbf{23.7} & \textbf{6.8} & \textbf{6.8} \\ \hline
\end{tabular}
\label{tab:results_fusion}
\end{table}
%%%%%%%%%%%%%%%%%%%%%%%%%%%%%%%%%%%%%

\subsection{Comparison with the State of the Art}
\label{sec:class_SOA}
Finally, we compare our results with the state of the art in~\tblref{tab:SOA}.
In the classification task of ILSVRC-2014 challenge~\citep{Russakovsky14}, our ``VGG'' team secured the 2nd place with $7.3\%$ test error using an ensemble
of 7 models. After the submission, we decreased the error rate to $6.8\%$ using an ensemble of 2 models.
% (which was achieved using a combination of 7 models as explained in~\sref{sec:fusion}).
% After the submission, we decreased the error rate to $7.0\%$ to by utilising two models, trained at multiple scales (as described above).

As can be seen from~\tblref{tab:SOA}, our very deep \mbox{ConvNets} significantly outperform the previous generation of models, which achieved the best results
in the ILSVRC-2012 and ILSVRC-2013 competitions. Our result is also competitive with respect to the classification task winner (GoogLeNet with $6.7\%$ error) and substantially
outperforms the ILSVRC-2013 winning submission Clarifai, which achieved $11.2\%$ with outside training data and $11.7\%$ without it.
This is remarkable, considering that our best result is achieved by combining just two models -- significantly less than used in most ILSVRC submissions. 
In terms of the single-net performance, our architecture achieves the best result ($7.0\%$ test error), outperforming a single GoogLeNet by $0.9\%$.
Notably, we did not depart from the classical ConvNet architecture of~\citet{LeCun89}, but improved it by substantially increasing the depth.
% We note that our best-performing single network was not submitted to the evaluation server before the ILSVRC deadline due to the lack of time.
% It was, however, used in our best-performing submissions based on the combination of multiple nets.

\vspace{-1em}

%%%%%%%%%%%%%%%%%%%%%%%%%%%%%%%%%%%%%
\begin{table}[htb]
\setlength{\tabcolsep}{2pt}
\small
\centering
\caption{\textbf{Comparison with the state of the art in ILSVRC classification}. Our method is denoted as ``VGG''.
Only the results obtained without outside training data are reported.
}
\begin{tabular}{|l|c|c|c|} \hline
Method & top-1 val.\ error (\%) & top-5 val.\ error (\%) & top-5 test error (\%) \\ \hline
VGG (2 nets, multi-crop \& dense eval.) & \textbf{23.7} & \textbf{6.8} & \textbf{6.8} \\ \hline
VGG (1 net, multi-crop \& dense eval.) & 24.4 & 7.1 & 7.0 \\ \hline\hline
% VGG (2 nets, dense eval.) & 24.0 & 7.1 & 7.0 \\ \hline
% VGG (1 net, dense eval.) & 24.8 & 7.5 & 7.3 \\ \hline\hline
VGG (ILSVRC submission, 7 nets, dense eval.) & 24.7 & 7.5 & 7.3 \\ \hline\hline
% VGG (ILSVRC submission, 1 net, dense eval.) & 24.9 & 8.0 & - \\ \hline\hline
GoogLeNet~\citep{Szegedy14} (1 net) & - & \multicolumn{2}{c|}{7.9} \\ \hline
GoogLeNet~\citep{Szegedy14} (7 nets) & - & \multicolumn{2}{c|}{\textbf{6.7}} \\ \hline
MSRA~\citep{He14} (11 nets) & - & - & 8.1 \\ \hline
MSRA~\citep{He14} (1 net) & 27.9 & 9.1 & 9.1 \\ \hline
Clarifai~\citep{Russakovsky14} (multiple nets) & -  & - & 11.7 \\ \hline
Clarifai~\citep{Russakovsky14} (1 net) & - & - & 12.5 \\ \hline
Zeiler \& Fergus~\citep{Zeiler13} (6 nets) & 36.0 & 14.7 & 14.8 \\ \hline
Zeiler \& Fergus~\citep{Zeiler13} (1 net) & 37.5 & 16.0 & 16.1 \\ \hline
OverFeat~\citep{Sermanet14} (7 nets) & 34.0 & 13.2 & 13.6 \\ \hline
OverFeat~\citep{Sermanet14} (1 net) & 35.7 & 14.2 & - \\ \hline
Krizhevsky~\etal~\citep{Krizhevsky12} (5 nets) & 38.1 & 16.4 & 16.4 \\ \hline
Krizhevsky~\etal~\citep{Krizhevsky12} (1 net) & 40.7 & 18.2 & - \\ \hline
\end{tabular}
\label{tab:SOA}
\end{table}
%%%%%%%%%%%%%%%%%%%%%%%%%%%%%%%%%%%%%

\vspace{-1em}

\section{Conclusion}
\label{sec:conclusion}
In this work we evaluated very deep convolutional networks (up to 19 weight layers) for large-scale image classification.
It was demonstrated that the representation depth is beneficial for the classification accuracy, and that 
state-of-the-art performance on the ImageNet challenge dataset can be achieved using a conventional ConvNet architecture~\citep{LeCun89,Krizhevsky12}
with substantially increased depth. 
% Namely, our object localisation system won the ILSVRC-2014 localisation challenge, while our classification system took the second place in the classification challenge.
In the appendix, we also show that our models generalise well to a wide range of tasks and datasets, matching or outperforming more complex recognition pipelines built around less deep image representations.
Our results yet again confirm the importance of depth in visual representations.

\subsubsection*{Acknowledgements}
This work was supported by ERC grant VisRec no.\ 228180. 
We gratefully acknowledge the support of NVIDIA Corporation with the donation of the GPUs used for this research.

\bibliographystyle{iclr2015}
{
    \small
    \setlength{\bibsep}{3pt}
    \bibliography{bib/shortstrings,bib/vgg_local,bib/vgg_other,bib/current}
}

\appendix

\section{Localisation}
\label{sec:loc}
In the main body of the paper we have considered the classification task of the ILSVRC challenge, and performed a thorough evaluation of ConvNet architectures of different depth.
In this section, we turn to the localisation task of the challenge, which we have won in 2014 with $25.3\%$ error.
It can be seen as a special case of object detection, where a single object bounding box should be predicted for each of the top-5 classes, irrespective of the actual number of objects of the class.
% State-of-the-art object detection systems~\citep{}
For this we adopt the approach of~\citet{Sermanet14}, the winners of the ILSVRC-2013 localisation challenge, with a few modifications.
Our method is described in~\sref{sec:loc_method} and evaluated in~\sref{sec:loc_eval}.

\subsection{Localisation ConvNet}
\label{sec:loc_method}
To perform object localisation, we use a very deep ConvNet, where the last fully connected layer predicts the bounding box location instead of the class scores.
A bounding box is represented by a 4-D vector storing its center coordinates, width, and height. There is a choice of whether the bounding box prediction is shared across all classes
(single-class regression, SCR~\citep{Sermanet14}) or is class-specific (per-class regression, PCR). In the former case, the last layer is 4-D, while in the latter it is 4000-D (since
there are 1000 classes in the dataset).
Apart from the last bounding box prediction layer, we use the ConvNet architecture D (\tblref{tab:config}), which contains 16 weight layers and was found to be the best-performing in the classification task
(\sref{sec:exp}). 
% This allows us to speed-up training by initialising with a pre-trained classification net.

\paragraph{Training.} 
Training of localisation ConvNets is similar to that of the classification ConvNets (\sref{sec:train}).
The main difference is that we replace the logistic regression objective with a Euclidean loss, which penalises the deviation of the predicted bounding box parameters from the ground-truth.
We trained two localisation models, each on a single scale: $S=256$ and $S=384$ (due to the time constraints, we did not use training scale jittering for our ILSVRC-2014 submission).
Training was initialised with the corresponding classification models (trained on the same scales), and the initial learning rate was set to $10^{-3}$.
We explored both fine-tuning all layers and fine-tuning only the first two fully-connected layers, as done in~\citep{Sermanet14}. The last fully-connected layer was
initialised randomly and trained from scratch.

\paragraph{Testing.} 
We consider two testing protocols. 
The first is used for comparing different network modifications on the validation set, and considers only the bounding box prediction for the ground truth class (to factor out the classification errors).
The bounding box is obtained by applying the network only to the central crop of the image.

The second, fully-fledged, testing procedure is based on the dense application of the localisation ConvNet to the whole image, similarly to the classification task (\sref{sec:test}).
The difference is that instead of the class score map, the output of the last fully-connected layer is a set of bounding box predictions.
To come up with the final prediction, we utilise the greedy merging procedure of~\citet{Sermanet14}, which first merges spatially close predictions (by averaging their coordinates), and then rates them
based on the class scores, obtained from the classification ConvNet. 
When several localisation ConvNets are used, we first take the union of their sets of bounding box predictions, and then run the merging procedure on the union.
We did not use the multiple pooling offsets technique of~\citet{Sermanet14}, which increases the spatial resolution of the bounding box predictions 
% by compensating for the loss of resolution after spatial pooling. 
and can further improve the results.

\subsection{Localisation Experiments}
\label{sec:loc_eval}

In this section we first determine the best-performing localisation setting (using the first test protocol),
and then evaluate it in a fully-fledged scenario (the second protocol).
% , where the labels are predicted by the classification \mbox{ConvNet}, and the bounding box prediction is done based on the whole image.
The localisation error is measured according to the ILSVRC criterion~\citep{Russakovsky14}, \ie the bounding box prediction is deemed correct if its intersection over union ratio with the ground-truth
bounding box is above $0.5$.

\paragraph{Settings comparison.}
As can be seen from~\tblref{tab:loc_comparison}, per-class regression (PCR) outperforms the class-agnostic single-class regression (SCR), which differs from the findings of~\citet{Sermanet14},
where PCR was outperformed by SCR. We also note that fine-tuning all layers for the localisation task leads to noticeably better results than fine-tuning only the fully-connected layers 
(as done in~\citep{Sermanet14}). In these experiments, the smallest images side was set to $S=384$; the results with $S=256$ exhibit the same behaviour and are not shown for brevity.

%%%%%%%%%%%%%%%%%%%%%%%%%%%%%%%%%%%%%%
\begin{table}[htb]
\setlength{\tabcolsep}{2pt}
\small
\centering
\caption{\textbf{Localisation error for different modifications} with
the simplified testing protocol: the bounding box is predicted from a single central image crop, and the ground-truth class is used.
All ConvNet layers (except for the last one) have the configuration D (\tblref{tab:config}), while
the last layer performs either single-class regression (SCR) or per-class regression (PCR). 
}
\begin{tabular}{|c|c|c|} \hline
Fine-tuned layers & regression type & GT class localisation error \\ \hline
\multirow{2}{*}{1st and 2nd FC} & SCR & 36.4 \\ \cline{2-3}
& PCR & 34.3 \\ \hline
all & PCR & \textbf{33.1} \\ \hline
\end{tabular}
\label{tab:loc_comparison}
\end{table}
%%%%%%%%%%%%%%%%%%%%%%%%%%%%%%%%%%%%%

\paragraph{Fully-fledged evaluation.}
Having determined the best localisation setting (PCR, fine-tuning of all layers), we now apply it in the fully-fledged scenario,
where the top-5 class labels are predicted using our best-performing classification system (\sref{sec:class_SOA}), and multiple densely-computed bounding box predictions are merged using 
the method of~\citet{Sermanet14}.
As can be seen from~\tblref{tab:loc_full}, application of the localisation ConvNet to the whole image substantially improves the results compared to using a center crop (\tblref{tab:loc_comparison}),
despite using the top-5 predicted class labels instead of the ground truth.
Similarly to the classification task (\sref{sec:exp}), testing at several scales and combining the predictions of multiple networks
further improves the performance.

%%%%%%%%%%%%%%%%%%%%%%%%%%%%%%%%%%%%%
\begin{table}[htb]
\small
\centering
\caption{\textbf{Localisation error}
}
\begin{tabular}{|c|c|c|c|} \hline
\multicolumn{2}{|c|}{smallest image side} & \multicolumn{2}{c|}{top-5 localisation error (\%)} \\ \hline
train ($S$) & test ($Q$) & val. & test. \\ \hline
256 & 256 & 29.5 & - \\ \hline
384 & 384 & 28.2 & 26.7\\ \hline
384 & 352,384 & 27.5 & - \\ \hline\hline
\multicolumn{2}{|c|}{fusion: 256/256 and 384/352,384} & \textbf{26.9} & \textbf{25.3} \\ \hline
\end{tabular}
\label{tab:loc_full}
\end{table}
% %%%%%%%%%%%%%%%%%%%%%%%%%%%%%%%%%%%%%

\paragraph{Comparison with the state of the art.}

We compare our best localisation result with the state of the art in~\tblref{tab:loc_SOA}.
With $25.3\%$ test error, our ``VGG'' team won the localisation challenge of ILSVRC-2014~\citep{Russakovsky14}.
Notably, our results are considerably better than those of the ILSVRC-2013 winner Overfeat~\citep{Sermanet14}, even though we used less scales and did not employ
their resolution enhancement technique. 
We envisage that better localisation performance can be achieved if this technique is incorporated into our method.
This indicates the performance advancement brought by our very deep ConvNets -- we got better results with a simpler localisation method, but a more powerful representation.

%%%%%%%%%%%%%%%%%%%%%%%%%%%%%%%%%%%%%
\begin{table}[htb]
\setlength{\tabcolsep}{2pt}
\small
\centering
\caption{\textbf{Comparison with the state of the art in ILSVRC localisation}. Our method is denoted as ``VGG''.
}
\begin{tabular}{|l|c|c|c|} \hline
Method & top-5 val.\ error (\%) & top-5 test error (\%) \\ \hline
VGG & \textbf{26.9} & \textbf{25.3} \\ \hline
GoogLeNet~\citep{Szegedy14} & - & 26.7 \\ \hline
OverFeat~\citep{Sermanet14} & 30.0 & 29.9 \\ \hline
Krizhevsky~\etal~\citep{Krizhevsky12} & - & 34.2 \\ \hline
\end{tabular}
\label{tab:loc_SOA}
\end{table}
%%%%%%%%%%%%%%%%%%%%%%%%%%%%%%%%%%%%%

\section{Generalisation of Very Deep Features}
\label{sec:dataset_transfer}
In the previous sections we have discussed training and evaluation of very deep ConvNets on the ILSVRC dataset. In this section, we evaluate our ConvNets, pre-trained on ILSVRC, as feature extractors on other, smaller, datasets, where training large models from scratch is not feasible due to over-fitting. Recently, there has been a lot of interest in such a use case~\citep{Zeiler13,Donahue13,Razavian14,Chatfield14}, as it turns out that deep image representations, learnt on ILSVRC, generalise well to other datasets, where they have outperformed hand-crafted representations by a large margin. Following that line of work, we investigate if our models lead to better performance than more shallow models utilised in the state-of-the-art methods.
% evaluated in the prior art~\citep{Zeiler13,Chatfield14}.
In this evaluation, we consider two models with the best classification performance on ILSVRC (\sref{sec:exp}) -- configurations ``Net-D'' and ``Net-E'' (which we made publicly available).

To utilise the ConvNets, pre-trained on ILSVRC, for image classification on other datasets, we remove the last fully-connected layer (which performs 1000-way ILSVRC classification), and use 4096-D activations of the penultimate layer as image features, which are aggregated across multiple locations and scales. The resulting image descriptor is $L_2$-normalised and combined with a linear SVM classifier, trained on the target dataset. For simplicity, pre-trained ConvNet weights are kept fixed (no fine-tuning is performed). 
% but we note that fine-tuning is beneficial for certain datasets~\citep{Chatfield14}.

Aggregation of features is carried out in a similar manner to our ILSVRC evaluation procedure (\sref{sec:test}). Namely, an image is first rescaled so that its smallest side equals $Q$, and then the network is densely applied over the image plane (which is possible when all weight layers are treated as convolutional). We then perform global average pooling on the resulting feature map, which produces a 4096-D image descriptor. The descriptor is then averaged with the descriptor of a horizontally flipped image.
As was shown in~\sref{sec:multi-scale}, evaluation over multiple scales is beneficial, so we extract features over several scales $Q$.
% , \eg $Q \in \{256,384,512\}$ (as used in ILSVRC experiments). 
The resulting multi-scale features can be either stacked or pooled across scales.
Stacking allows a subsequent classifier to learn how to optimally combine image statistics over a range of scales; this, however, comes at the cost of the increased descriptor dimensionality.
We return to the discussion of this design choice in the experiments below.
We also assess late fusion of features, computed using two networks, which is performed by stacking their respective image descriptors.

% Interestingly, it was found that stacking performs better than averaging on Caltech-101 and Caltech-256, while on VOC-2007 and VOC-2012 averaging performs slightly better (and 
% A potential reason for that difference is that in VOC images the objects can appear over a variety of scales, so there is no particular semantics associated with each scale, and average pooling across scales is meaningful. In Caltech images, however, objects normally occupy the whole image, so multi-scale image features are semantically different (capturing the whole object or object parts), and stacking (rather than averaging) allows a classifier to exploit scale-specific representations.

%%%%%%%%%%%%%%%%%%%%%%%%%%%%%%%%%%%%%
\begin{table}[htb]
\setlength{\tabcolsep}{2pt}
\small
\centering
\caption{\textbf{Comparison with the state of the art in image classification on VOC-2007, VOC-2012, Caltech-101, and Caltech-256}.
Our models are denoted as ``VGG''. 
Results marked with * were achieved using \mbox{ConvNets} pre-trained on the \emph{extended} ILSVRC dataset (2000 classes).
}
\begin{tabular}{|l|c|c|c|c|} \hline
\multirow{2}{*}{Method} & VOC-2007 & VOC-2012 & Caltech-101 & Caltech-256  \\ 
& (mean AP) & (mean AP) & (mean class recall) & (mean class recall) \\ \hline
Zeiler \& Fergus~\citep{Zeiler13} & - & 79.0 & 86.5 $\pm$ 0.5 & 74.2 $\pm$ 0.3 \\ 
Chatfield~\etal~\citep{Chatfield14} & 82.4 & 83.2 & 88.4 $\pm$ 0.6 & 77.6 $\pm$ 0.1 \\ 
He~\etal~\citep{He14} & 82.4 & - & \textbf{93.4 $\pm$ 0.5} & - \\ 
Wei~\etal~\citep{Wei14} & 81.5 (85.2$^*$) & 81.7 (\textbf{90.3$^*$}) & - & - \\ \hline\hline
VGG Net-D (16 layers) & 89.3 & 89.0 & 91.8 $\pm$ 1.0 & 85.0 $\pm$ 0.2 \\ 
VGG Net-E (19 layers) & 89.3 & 89.0 & 92.3 $\pm$ 0.5 & 85.1 $\pm$ 0.3 \\
VGG Net-D \& Net-E & \textbf{89.7} & \textbf{89.3} & 92.7 $\pm$ 0.5 & \textbf{86.2 $\pm$ 0.3} \\ \hline
\end{tabular}
\label{tab:generalise_class}
\end{table}
%%%%%%%%%%%%%%%%%%%%%%%%%%%%%%%%%%%%%

\paragraph{Image Classification on VOC-2007 and VOC-2012.}
We begin with the evaluation on the image classification task of PASCAL VOC-2007 and VOC-2012 benchmarks~\citep{Everingham15}. These datasets contain 10K and 22.5K images respectively, and each image is annotated with one or several labels, corresponding to 20 object categories. The VOC organisers provide a pre-defined split into training, validation, and test data (the test data for VOC-2012 is not publicly available; instead, an official evaluation server is provided). Recognition performance is measured using mean average precision (mAP) across classes.
% TODO: explain why mAP for ICLR - since VOC is a multi-label dataset
% Due to the multi-label image annotation, we trained an ensemble of binary one-vs-rest linear SVM classifiers using LIBLINEAR~\citep{Fan08}. 

Notably, by examining the performance on the validation sets of \mbox{VOC-2007} and VOC-2012, we found that aggregating image descriptors, computed at multiple scales, by averaging performs similarly to the aggregation by stacking.
% (max-pooling over scales performed slightly worse). 
We hypothesize that this is due to the fact that in the VOC dataset the objects appear over a variety of scales, so there is no particular scale-specific semantics which a classifier could exploit.
Since averaging has a benefit of not inflating the descriptor dimensionality, we were able to aggregated image descriptors over a wide range of scales: $Q \in \{256,384,512,640,768\}$. It is worth noting though that the improvement over a smaller range of $\{256,384,512\}$ was rather marginal ($0.3\%$).

The test set performance is reported and compared with other approaches in~\tblref{tab:generalise_class}. Our networks ``Net-D'' and ``Net-E'' exhibit identical performance on VOC datasets, and their combination slightly improves the results. Our methods set the new state of the art across image representations, pre-trained on the ILSVRC dataset, outperforming the previous best result of~\citet{Chatfield14} by more than $6\%$. It should be noted that the method of~\citet{Wei14}, which achieves $1\%$ better mAP on VOC-2012, is pre-trained on an extended 2000-class ILSVRC dataset, which includes additional 1000 categories, semantically close to those in VOC datasets. It also benefits from the fusion with an object detection-assisted classification pipeline.

\paragraph{Image Classification on Caltech-101 and Caltech-256.}
In this section we evaluate very deep features on \mbox{Caltech-101}~\citep{FeiFei04} and \mbox{Caltech-256}~\citep{Griffin07} image classification benchmarks. Caltech-101 contains 9K images labelled into 102 classes (101 object categories and a background class), while Caltech-256 is larger with 31K images and 257 classes. 
A standard evaluation protocol on these datasets is to generate several random splits into training and test data and report the average recognition performance across the splits, which is measured by the mean class recall (which compensates for a different number of test images per class).
Following~\citet{Chatfield14,Zeiler13,He14}, on Caltech-101 we generated 3 random splits into training and test data, so that each split contains 30 training images per class, and up to 50 test images per class. On Caltech-256 we also generated 3 splits, each of which contains 60 training images per class (and the rest is used for testing). In each split, 20\% of training images were used as a validation set for hyper-parameter selection.
% Unlike VOC, each image in Caltech-101/256 is annotated with a single label, so we utilised multi-class linear SVM formulation of~\citep{Crammer01}, implemented in LIBLINEAR.

We found that unlike VOC, on Caltech datasets the stacking of descriptors, computed over multiple scales, performs better than averaging or max-pooling. 
This can be explained by the fact that in Caltech images objects typically occupy the whole image, so multi-scale image features are semantically different (capturing the whole object \vs object parts), and stacking allows a classifier to exploit such scale-specific representations. We used three scales $Q \in \{256,384,512\}$.

Our models are compared to each other and the state of the art in~\tblref{tab:generalise_class}. 
As can be seen, the deeper 19-layer Net-E performs better than the 16-layer Net-D, and their combination further improves the performance.
On Caltech-101, our representations are competitive with the approach of~\citet{He14}, which, however, performs significantly worse than our nets on VOC-2007. On Caltech-256, our features outperform the state of the art~\citep{Chatfield14} by a large margin ($8.6\%$). 

\paragraph{Action Classification on VOC-2012.}
We also evaluated our best-performing image representation (the stacking of Net-D and Net-E features) on the PASCAL VOC-2012 action classification task~\citep{Everingham15}, which consists in predicting an action class from a single image, given a bounding box of the person performing the action. The dataset contains 4.6K training images, labelled into 11 classes. Similarly to the VOC-2012 object classification task, the performance is measured using the mAP. We considered two training settings: 
(i) computing the ConvNet features on the whole image and ignoring the provided bounding box;
(ii) computing the features on the whole image and on the provided bounding box, and stacking them to obtain the final representation.
The results are compared to other approaches in~\tblref{tab:generalise_action}.
%%%%%%%%%%%%%%%%%%%%%%%%%%%%%%%%%%%%%
\begin{table}[htb]
\setlength{\tabcolsep}{2pt}
\small
\centering
\caption{\textbf{Comparison with the state of the art in single-image action classification on VOC-2012}.
Our models are denoted as ``VGG''. 
Results marked with * were achieved using \mbox{ConvNets} pre-trained on the \emph{extended} ILSVRC dataset (1512 classes).
}
\begin{tabular}{|l|c|} \hline
Method & VOC-2012 (mean AP) \\ \hline
\citep{Oquab14} & 70.2$^*$ \\ \hline
\citep{Gkioxari14} & 73.6 \\ \hline
\citep{Hoai14a} & 76.3 \\ \hline\hline
VGG Net-D \& Net-E, image-only & \textbf{79.2} \\ \hline
VGG Net-D \& Net-E, image and bounding box & \textbf{84.0} \\ \hline
\end{tabular}
\label{tab:generalise_action}
\end{table}
%%%%%%%%%%%%%%%%%%%%%%%%%%%%%%%%%%%%%

Our representation achieves the state of art on the VOC action classification task even without using the provided bounding boxes, and the results are further improved when using both images and bounding boxes.
Unlike other approaches, we did not incorporate any task-specific heuristics, but relied on the representation power of very deep convolutional features.

\paragraph{Other Recognition Tasks.}
Since the public release of our models, they have been actively used by the research community for a wide range of image recognition tasks, consistently outperforming more shallow representations. For instance, \citet{Girshick14a} achieve the state of the object detection results by replacing the ConvNet of~\citet{Krizhevsky12} with our 16-layer model. Similar gains over a more shallow architecture of~\citet{Krizhevsky12} have been observed in semantic segmentation~\citep{Long14}, image caption generation~\citep{Kiros14,Karpathy14a}, texture and material recognition~\citep{Cimpoi14a,Bell14}.

%\paragraph{Summary.}
%Our experiments on PASCAL VOC and Caltech image classification datasets demonstrate that a better performance of very deep representations on ILSVRC translates into a better %performance on other datasets. We note that our results are consistently high across different datasets despite simple training and evaluation procedure.

\section{Paper Revisions}
\label{sec:revisions}
Here we present the list of major paper revisions, outlining the substantial changes for the convenience of the reader.

\textbf{v1} Initial version. Presents the experiments carried out before the ILSVRC submission.

\textbf{v2} Adds post-submission ILSVRC experiments with training set augmentation using scale jittering, which improves the performance.

\textbf{v3} Adds generalisation experiments (\apref{sec:dataset_transfer}) on PASCAL VOC and Caltech image classification datasets. The models used for these experiments are publicly available.

\textbf{v4} The paper is converted to ICLR-2015 submission format. Also adds experiments with multiple crops for classification.

\textbf{v6} Camera-ready ICLR-2015 conference paper. Adds a comparison of the net B with a shallow net and the results on PASCAL VOC action classification benchmark.

\end{document}